# Effective Quantization Approaches for Recurrent Neural Networks

Md Zahangir Alom[1], Adam T Moody[2], Naoya Maruyama[2], Brian C Van Essen[2], and Tarek M. Taha[1]
[1]Department of Electrical and Computer Engineering, University of Dayton, OH 45469, USA.
[2]Center for Applied Scientific Computing, Lawrence Livermore National Laboratory, CA 94550, USA.
e-mail: [1]{alomm1, tahta1}@udayton.edu, [2]{moody20, nmaruyama, vanessen1}@llnl.gov

*Abstract*— **Deep learning, Recurrent Neural Networks (RNN) in particular have shown superior accuracy in a large variety of tasks including machine translation, language understanding, and movie frames generation. However, these deep learning approaches are very expensive in terms of computation. In most cases, Graphic Processing Units (GPUs) are in used for large scale implementations. Meanwhile, energy efficient RNN approaches are proposed for deploying solutions on special purpose hardware including Field Programming Gate Arrays (FPGAs) and mobile platforms. In this paper, we propose an effective quantization approach for Recurrent Neural Networks (RNN) techniques including Long Short Term Memory (LSTM), Gated Recurrent Units (GRU), and Convolutional Long Short Term Memory (ConvLSTM). We have implemented different quantization methods including Binary Connect {-1, 1}, Ternary Connect {-1, 0, 1}, and Quaternary Connect {-1, -0.5, 0.5, 1}. These proposed approaches are evaluated on different datasets for sentiment analysis on IMDB and video frame predictions on the moving MNIST dataset. The experimental results are compared against the full precision versions of the LSTM, GRU, and ConvLSTM. They show promising results for both sentiment analysis and video frame prediction.**

## I. INTRODUCTION

Deep Neural Networks have been successfully applied and have achieved superior recognition accuracies in different application domains such as computer vision, speech processing, natural language processing (NLP), and medical imaging [1,2]. Several variants of deep learning approaches have been trained and tested with deeper and wider networks for achieving classification accuracies which are similar to, or sometimes beyond, human level recognition accuracies. Typically, when size of a neural network increases, it becomes more powerful and provides better classification accuracies. This comes at the significantly increasing costs of storage consumption, memory bandwidth, and computational cost. In most of the cases, the training is being executed on GPUs for dealing with big data volumes. This is very expensive in terms of power. In addition, deep learning approaches are expensive in terms of the number of networks parameters. This requires large storage and runtime memory for use. On the other hand, these types of massive scale implementations with large numbers of network parameters are not suitable for low power implementation, such as, unmanned aerial vehicles (UAV), medical devices, low memory system such as mobile devices, and Field Programmable Gate Arrays (FPGA).

Several research efforts are on-going to develop better networks with lower computation costs and fewer network parameters for low-power and low-memory systems without dropping classification accuracy. There are two main ways to design very efficient deep network structures: the first approach is by optimizing the internal operational cost with efficient network architectures. The second approach is to design networks with low precision operations for hardware efficient networks. As far as the network structure is concerned, the number of parameters can be reduced dramatically by using low dimensional convolutional filters in the convolutional layer as this also helps to add more non-linearity to networks [3,4]. One intuition is that larger activation maps (due to delayed down-sampling) can lead to higher classification accuracies [3]. This intuition has been investigated by K. He and H. Sun by applying delayed down-sampling into four different architectures of CNNs. It was observed that in each case, delayed down-sampling led to higher classification accuracies [5].

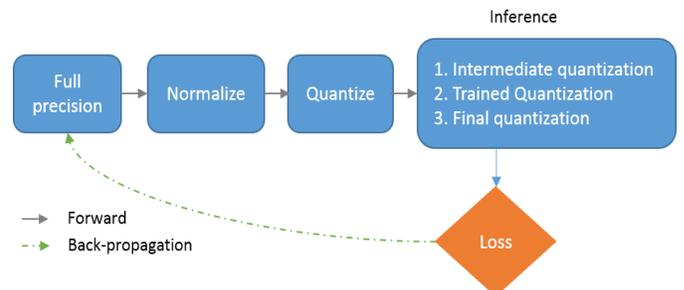

**Fig. 1**. Quantization approach of a deep neural networks [26].

Computation cost and memory can be saved significantly with lower precision multiplications and fewer multiplications through drop connection [6, 7]. These papers introduced Binary Connect Neural Networks (BNN) and Ternary Connect Neural Networks (TNN). Generally, multiplication of a real-valued weight by a real-valued activation (in the forward propagations) and gradient calculation (in the backward propagations) are the main operations of deep neural networks. Binary connect or BNN is a technique that eliminates the multiplication operations by converting the weights used in the forward propagation to be binary, i.e. constrained to only two values (0 and 1 or -1 and 1). As a result, the multiplication operations can be performed with simple additions (and subtractions), making the training process faster. There are two ways to represent real values to its corresponding binary values: deterministic and stochastic. In the deterministic technique, a straightforward thresholding technique is applied to the weights. In the stochastic approach, a matrix is converted to binary based on probabilities where the *"hard sigmoid"* function is used because it is computationally inexpensive. Experimental result show significantly better recognition performance on different benchmarks, including ImageNet [8, 9, 10]. A flow diagram of the quantization approach is shown in Fig. 1, based on the recently published paper [26]. There are several advantages of BNNs: first, it is observed that binary multiplications on GPUs are almost seven times faster than traditional matrix multiplications on GPU. Second, in the forward pass, BNNs drastically reduce memory size and accesses, and replace most arithmetic operations with bit-wise operations, which leads to great increases of power efficiency. Third, binarized kernels can be used in CNNs, which can reduce the complexity of dedicated hardware by 60%. Forth, it is also observed that memory accesses typically consume more energy compare to arithmetic operations and that memory access costs increase with memory size. BNNs are beneficial with respect to both aspects.

Other techniques have also been proposed in the last few years [11, 12, 13]. Another power efficient and hardware friendly network structure has been proposed for CNNs with XNOR operations. In XNOR based CNN implementations, both the filters and inputs to the convolution layer are binary. This results in about 58x faster convolutional operations and 32x memory savings. In the same paper, Binary Weight Networks (BWN) have been proposed, which enable around 32x memory savings, allowing implementation of state-of-the-art networks on CPUs for real time operations instead of GPUs. This model was tested on the ImageNet dataset and provided only 2.9% less classification accuracy than the full-precision AlexNet (in the top-1% measure). This network required less power and computation time. It accelerated the training process of deep neural networks dramatically for specialized hardware implementations [14]. The Energy Efficient Deep Neural Network (EEDN) architecture was first proposed for neuromorphic systems in 2016. In addition, they released a deep learning framework called EEDN, which provides accuracies that are very close to the state-of-the art for almost all the popular benchmarks except the ImageNet dataset [15,16].

Some papers have been published recently which are based on quantization approaches proposed for RNNs [17, 18, 19]. However, in this paper, we have proposed effective quantization methods for RNN and empirically evaluated the performance on different datasets. The contribution of this work can be summarized as follows:

- Proposed effective quantization with binary connect, ternary connect, and quaternary connect approaches for RNNs.
- Evaluated on three different recurrent methods including LSTM, GRU, and ConvLSTM.
- Performance evaluation of LSTM and GRU for sentiment analysis on amazon IMDB dataset.
- To our knowledge, first time toward the evaluation of quantized ConvLSTM for video frame generation with the moving MNIST dataset.

These efficient proposed quantization approaches will help to implement power efficient Deep Learning (DL) on FPGAs and embedded devices, including low power mobile devices. The paper is organized as follows: Section II discusses recurrent neural networks (RNN). The proposed method in detail is demonstrated in Section III. Section IV, explains datasets, experiments, and results. Conclusions and future directions are given in Section V.

## II. RECURRENT NEURAL NETWORKS (RNN)

Human thoughts have persistence; Humans don't throw everything away and start their thinking from scratch every second. When you are reading a novel, you are understanding each word or sentence based on the understanding of previous words or sentences. The traditional neural network approaches including DNN and CNN cannot deal with this type of problems. The standard Neural Networks and CNNs are incapable of this due to the following reasons. First, these approaches only handle a fixed-size vector as input (e.g., an image or video frame) and produce a fixed-size vector as output (e.g., probabilities of different classes). Second, those models operate with a fixed number of computational steps (e.g. the number of layers in the model). The RNNs are unique as they allow operation over a sequence of vectors over time. A very basic RNN model, where the outputs from the hidden layers are used as inputs with the inputs of hidden layers [20] is

$$h_t = \sigma_h(w_h x_t + u_h h_{t-1} + b_h) \quad (1)$$

$$y_t = \sigma_y(w_y h_t + b_y) \quad (2)$$

where $x_t$ is the input vector, $h_t$ is the hidden layer vectors, $y_t$ is the output vector, *w and u* are weight matrices, and *b* is the bias vector. A loop allows information to be passed from one step of the network to the next. A RNN can be thought of as multiple copies of the same network, each network passing a message to a successor. The diagram below shows what happens if we unroll the loop of a RNN model.

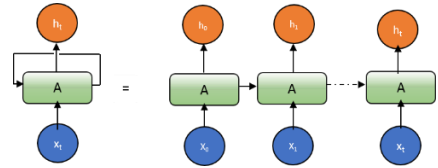

**Fig. 2.** An unrolled RNNs.

The main problem is vanishing gradient problem to learn RNN approach depending upon the length of input sequences. For the very first time, this problem is solved by Hochreiter el at. in 1992 [21]. However, there are several solutions that have been proposed for solving the vanishing gradient problem of RNN approaches in recent decades. Two possible effective solutions of this problem are: first, clip the gradient (scale the gradient if its norm is too big) and second, better RNN models.

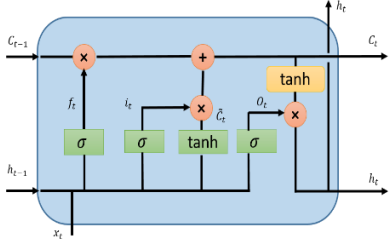

**Fig. 3.** Diagram for Long Short Term Memory (LSTM).

*A. Long Short Term Memory (LSTM)*

One of the improved models is introduced by Felix A. el at. in 2000, which is known as Long Short-Term Memory (LSTM) [22]. From then on, there are different variants of models that have been proposed based on this model. This improved version of RNN approaches allow larger sequences in the input, the output, or in the most general case, both and applying vastly for text mining, language understanding efficiently. The key idea of LSTMs is the cell state, the horizontal line running through the top of the Fig. 3. LSTM removes or adds information to the cell state called gates: input gate($i_t$), forget gate ($f_t$), and output gate($o_t$) can be defined as:

$$f_t = \sigma(W_f.[h_{t-1}, x_t] + b_f) \quad (4)$$

$$i_t = \sigma(W_i.[h_{t-1}, x_t] + b_i) \quad (5)$$

$$\tilde{C}_t = tanh(W_C.[h_{C-1}, x_t] + b_C) \quad (6)$$

$$C_t = f_t * C_{t-1} + i_t * \tilde{C}_t \quad (7)$$

$$O_t = \sigma(W_O.[h_{t-1}, x_t] + b_O) \quad (8)$$

$$h_t = O_t * tanh(C_t) \quad (9)$$

The LSTM model is very popular for temporal information processing. Most of the paper includes LSTM model with some variant, which is very minor.

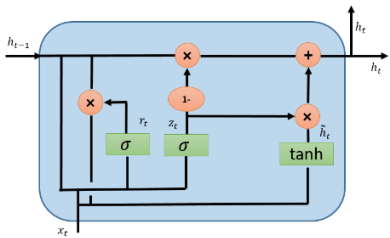

**Fig. 4.** Diagram for Gated Recurrent Unit (GRU).

*B. Gated Recurrent Unit (GRU)*

GRU also from LSTM with a slightly more variation by Cho, et al. in 2014, which is now very popular in the community working with recurrent networks. The main reason of the popularity is computation cost and simplicity of the model, which is shown in Fig. 4. GRU is a significantly lighter version of RNN approach than standard LSTM in term of topology, computation cost, and complexity [23]. This technique is combined with the forget and input gates into a single "update gate" and merges the cell state and hidden state, and makes some other changes. The simpler model of GRU has been growing increasingly popular. Mathematically GRU can be expressed with the following equations:

$$z_t = \sigma(W_z.[h_{t-1}, x_t]) \quad (10)$$

$$r_t = \sigma(W_r.[h_{t-1}, x_t]) \quad (11)$$

$$\tilde{h}_t = tanh(W.[r_t * h_{t-1}, x_t]) \quad (12)$$

$$h_t = (1 - z_t) * h_{t-1} + z_t * \tilde{h}_t \quad (13)$$

According to different empirical studies, there is no clear evidence of the winner. However, GRU requires fewer network parameters, which makes the model faster. On the other hand, LSTM provides better performance, if you have enough data and computational power [24].

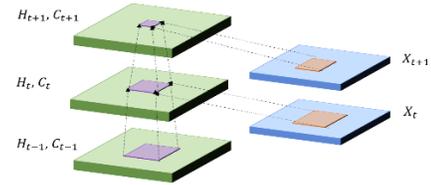

**Fig. 5.** Pictorial diagram for ConvLSTM unit [11].

In this work, we have evaluated both quantized version of LSTM and GRU for sentiment analysis in this implementation.

*C. Convolutional LSTM (ConvLSTM)*

The problem with fully connected (FC) LSTM in short FC-LSTM model is handling spatiotemporal data and its usage of full connection in input-to-state and state-to-state transactions, where no spatial information has been encoded. In ConvLSTM model, the internal gates of ConvLSTM are 3D tensor, where last two dimensions are spatial dimensions (rows and columns). The ConvLSTM determines the future states of a certain cell in the grid with respect to inputs and the past states of its local neighbors which can be achieved using convolution operation in the state-to-state or inputs-to-states transition show in Fig. 5.

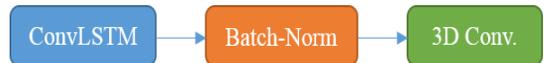

**Fig. 6.** ConvLSTM layer with batch-normalization and 3D convolution.

ConvLSTM provides very good performance for temporal data analysis with video dataset [11]. Mathematically, the ConvLSTM is expressed as follows, where * represents the convolution operation and ∘ denotes for Hadamard product:

$$i_t = \sigma(w_{xi}.\mathcal{X}_t + w_{hi} * \mathcal{H}_{t-1} + w_{hi} \circ \mathcal{C}_{t-1} + b_i) \quad (14)$$

$$f_t = \sigma(w_{xf}.\mathcal{X}_t + w_{hf} * \mathcal{H}_{t-1} + w_{hf} \circ C_{t-1} + b_f) \quad (15)$$

$$\widetilde{C}_t = \tanh(w_{xc}.\mathcal{X}_t + w_{hc} * \mathcal{H}_{t-1} + b_C) \quad (16)$$

$$C_t = f_t \circ C_{t-1} + i_t * \widetilde{C}_t \quad (17)$$

$$o_t = \sigma(w_{xo}.\mathcal{X}_t + w_{ho} * \mathcal{H}_{t-1} + w_{ho} \circ C_t + b_o) \quad (18)$$

$$h_t = o_t \circ \tanh(C_t) \quad (19)$$

In this implementation, we have used a very basic ConvLSTM structure where a single ConvLSTM layer, one batch-norm layer, and one 3D reconstruction layer are used. The basic diagram is shown in Fig. 6.

### III. PROPOSED QUANTIZATION APPROACHES

To quantize of the weights of a neural network, the quantization techniques are applied in the forward propagation, which reduces the operations compared to full precision and reduces memory requirement significantly. After calculating the loss of the model, weight gradients are updated with respect to the full precision weight values. The flow diagram according to the Ternary connect neural networks [26] is shown in Fig. 1. According to the ternary connect quantization method, the value of $\pm\Delta$ is optimized by minimizing the expectation of $l_2$ distance between full precision and ternary weights. The maximum absolute value of the weights is used as a reference threshold to the layers and maintain a constant factor $t$ for all the layers, which represents with $\Delta_l = t \times \max(|\widetilde{w}|)$. They maintain a constant sparsity $r$ for all layers throughout training and this hyper parameter $r$ helps to obtain ternary weight networks with various sparsities. $t = 0.05$ is used in the experiment of CIFAR 10 and ImageNet dataset [26].

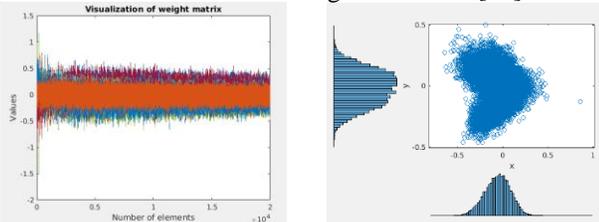

**Fig. 7.** Visualization of weights on the left and weight distribution is shown on the right.

In addition, another very close work for ternary connect networks, the weights ($W$) are uniformly or normal distributed in $[-a, a]$ and $\Delta$ lies in $[0, a]$. In case of uniform distribution: the approximated value is $\frac{1}{3}a$, which is equal to $\frac{2}{3}E(|W|)$. For normal distribution, $N(0, \sigma^2)$, the approximated $\Delta^*$ is $\frac{1}{3}\sigma$, which is equal to $0.75 * E(|W|)$ is used. Finally, this paper proposed a rule of thumb that $\Delta^* = 0.75 * E(|W|) \approx \frac{0.7}{n}\sum_{i=1}^{n}|W_i|$, which is a strictly optimized threshold [27]. Furthermore, according to [17], the weights follow the characteristics of normal distribution and therefore they assume $W$ has a symmetric distribution around zero. They scaled the mean absolute weights with a factor of 0.25 and evaluated for different bits for weights and activation. A straight-through estimator is used for this implementation [17].

Like others, we have determined the threshold values with basic statistics (mean and standard deviation) of weights in a layer. However, if we observe Fig. 7, it shows that the weight distribution is normal with mean ($\mu$) and the standard deviation ($\sigma$). In addition, we observe that most of the weights values fall very close to zero. For binary connect neural networks, the thresholding is done with respect to zero on normalized weights of a layer. The equation is as follows:

$$\begin{array}{ll} if\ w_0 \geq 0 & 1 \\ otherwise & -1 \end{array} \quad (20)$$

Fig. 8 shows the outputs distribution of weights after applying Eq. 20. It demonstrates clearly that the weights in a layer are uniformly distributed with respect to zero. Thus, we do not need to worry about the distribution of quantized weights for BC.

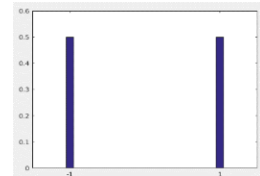

**Fig. 8.** Distribution of weight for binary connect.

However, the ternary connect neural network contains the values of {-1, 0, 1}. In ternary connect networks, we have calculated the mean ($\mu$) standard deviation ($\sigma$) of the weight of a layer. To achieve the approximate normal distribution of the quantize weights, the following equation is applied:

$$\begin{array}{ll} if\ w \leq -(\mu + \sigma) & -1 \\ -(\mu + \sigma) < w \leq (\mu + \sigma) & 0 \\ w > (\mu + \sigma) & 1 \end{array} \quad (21)$$

After applying the quantization with Eq. 22, the resulting quantized weights show approximated normal distribution which is stated in Fig 9(b). However, if we apply Eq. 22 then we achieve uniform distribution for the quantized weights.

$$\begin{array}{ll} if\ w \leq -\left(\mu + \frac{\sigma}{2}\right) & -1 \\ -\left(\mu + \frac{\sigma}{2}\right) < w \leq \left(\mu + \frac{\sigma}{2}\right) & 0 \\ w > \left(\mu + \frac{\sigma}{2}\right) & 1 \end{array} \quad (22)$$

The following figure shows normal and uniform distribution graphically in Fig 9(b) and Fig. 9(c) respectively.

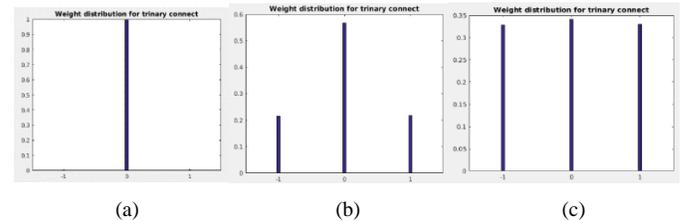

**Fig. 9.** Weight distribution for ternary: (a) {-0.5, 0.5} (b) Weight distribution for Eq. 21, and (c) Weight distribution for Eq. 22 as threshold.

From Fig. 9 (a), if we use {-0.5, 0.5} as threshold like other approaches, then the resulting weight distribution is unary where almost all weights get a single value of zero. However, we have applied thresholding according to the Eq. 21 and 22. This proposed approach ensures proper weight distributions shown in Fig. 9(b) and Fig. 9(c) respectively. In the QC approach, the {-1, -0.5, 0.5, 1} values are considered for weight representation. The threshold values are determined based on Eq. 23 and Eq. 24, which produces normal and uniform distribution of quantized weights.

$$\begin{cases} if\ w \leq -\left(\mu + \frac{\sigma}{4}\right) & -1 \\ -\left(\mu + \frac{\sigma}{4}\right) < w \leq 0 & -0.5 \\ 0 < w \leq \left(\mu + \frac{\sigma}{4}\right) & 0.5 \\ w > \left(\mu + \frac{\sigma}{4}\right) & 1 \end{cases} \quad (23)$$

To implement the uniform distribution of weights after quantization

$$\begin{cases} if\ w \leq -\left(\mu + \frac{\sigma}{6}\right) & -1 \\ -\left(\mu + \frac{\sigma}{6}\right) < w \leq 0 & -0.5 \\ 0 < w \leq \left(\mu + \frac{\sigma}{6}\right) & 0.5 \\ w > \left(\mu + \frac{\sigma}{6}\right) & 1 \end{cases} \quad (24)$$

The visualization of weights distribution after applying different thresholding are shown in Fig. 10. From the figure, it can be clearly observed that if we apply {-0.5, 0, 0.5} as threshold then instead of quaternary connected, it works like binary connect network with values of {-0.5, 0.5}, which is shown in Fig. 10 (a). However, the proposed approach shows proper normal and uniform distribution of quantized weights which is shown in Fig.10 (b) and Fig. 10 (c) respectively.

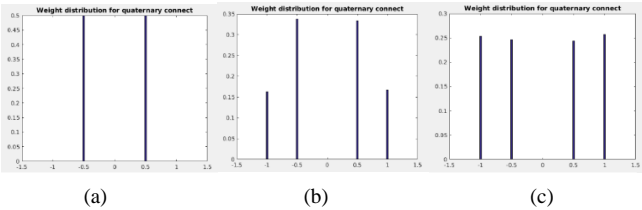

**Fig. 10**. Weight distribution for quaternary: (a) {- 0.5, 0, 0.5} (b) after applying Eq. 23 and (c) Outputs for Eq. 24 as threshold.

IV. RESULTS AND DISCUSSION

The entire experiment has been conducted in the Surface cluster of the Supercomputing center at the Lawrence Livermore National Laboratory (LLNL) and is implemented with Keras and TensorFlow. We have evaluated our proposed quantization techniques for sentiment analysis on the IMDB dataset [28] and movie frame prediction task on the moving MNIST dataset [29]. Before going to the main experiment, we have experimented on a very simple summation problem for selecting appropriate weight distributions. We have evaluated the full precision (FP) and three approaches with quantization including Binary Connect (BC), Ternary Connect (TC), and Quaternary Connect (QC). The inputs set contains 12 characters including {'0', '1', '2', '3', '4', '5', '6', '7', '8', '9', '+', ' '}. We have encoded each character with a binary value which is shown with orange color in Fig. 11. In the testing phase, after getting the encoding outputs shown in blue, we have decoded values for producing the desired outputs.

**Fig. 11.** Inputs, encoding system for summation problem.

[3, 10]] [13]
[' 3+10'] ['13']
[[11, 3, 10, 1, 0]] [[1, 3]]
[[[0, 0, 0, 0, 0, 0, 0, 0, 0, 0, 0, 1],
[0, 0, 0, 1, 0, 0, 0, 0, 0, 0, 0, 0],
[0, 0, 0, 0, 0, 0, 0, 0, 0, 0, 1, 0],
[0, 1, 0, 0, 0, 0, 0, 0, 0, 0, 0, 0],
[1, 0, 0, 0, 0, 0, 0, 0, 0, 0, 0, 0]]]
[[[0, 1, 0, 0, 0, 0, 0, 0, 0, 0, 0, 0],
[0, 0, 0, 1, 0, 0, 0, 0, 0, 0, 0, 0]]]

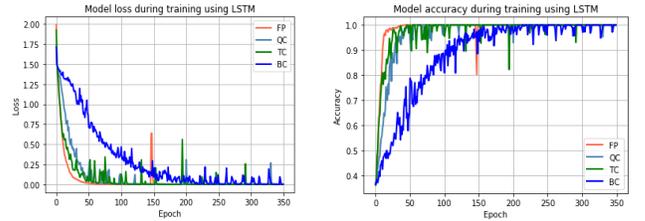

**Fig. 12.** Model loss on the left and accuracy on the right for LSTM

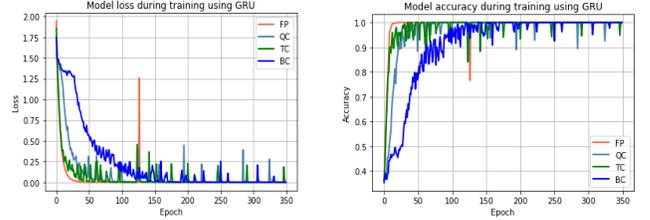

**Fig. 13.** Model loss and accuracy for GRU are shown on the left and right side respectively.

Inputs in the first two rows and third row shows encoding position. For example, before 3 there is a space. Encoding position number is 11. Orange color represents the encoded vectors of inputs. The blue color shows the encoded outputs which is equivalent of 13. We have experimented for normal distribution (ND) and uniform distribution (ED) of weights after quantization for LSTM and GRU. Total 1000 samples per epoch are considered and this experiment is run for 350 epochs shown in Fig. 11.

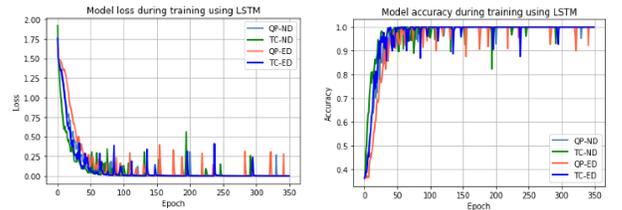

**Fig. 14**. Loss for ND and ED using LSTM on the left and accuracy on the right.

Fig. 12 and 13 show the training loss and accuracy for LSTM and GRU for the summation problem respectively. From Fig. 12, it can be observed that the LSTM and GRU with full precision show better performance than other quantization approaches. The same behavior is observed for accuracy as well. It is also noticed that the TC and QC version of LSTM and GRU provides promising training accuracy compared to the of LSTM and GRU with full precision. Fig. 14 shows the loss and accuracy of LSTM for normal and uniform distribution respectively. From the figure, it can be clearly observed that the approximate normal distribution performs better than the uniform distribution. Thus, the approximate normal distribution is used for TC and QC for the following experiments. The results are compared against the performance of LSTM, GRU, and ConvLSTM with full precision (32 bits) for all datasets. In this experiment, we have used ADAM optimizer and binary cross entropy loss.

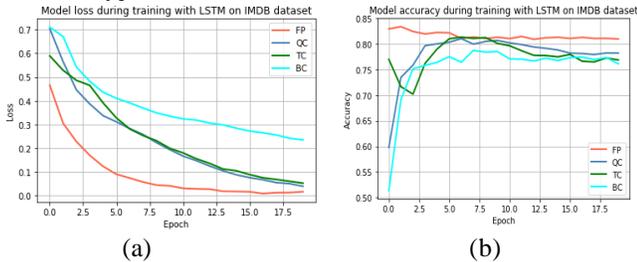

(a)            (b)

**Fig. 15**. Model loss and accuracy during training for LSTM. (a) Loss and (b) Accuracy.

### A. Sentiment Analysis:

The experiment is conducted on sequence to sequence problems for addition and IMDB sentiment analysis datasets. Here we report preliminary results that demonstrate the effectiveness of the proposed quantization methods on learning of recurrent models of LSTM and GRU. To accomplish this, the IMDB sentiment analysis dataset is used with max-feature numbers 20000, max number of words 80, and batch size 64. In both LSTM and GRU architectures, we have considered hidden units 128, and number of epochs 20.

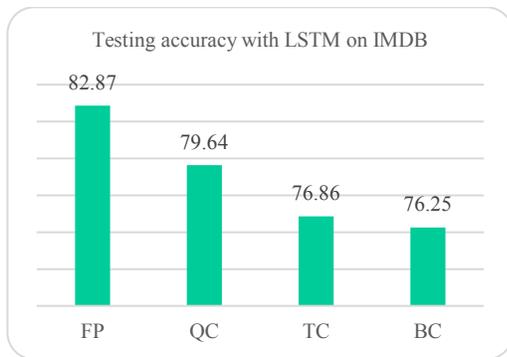

**Fig. 16**. Testing accuracy using LSTM.

#### 1) Results with LSTM
The training loss and validation accuracy with LSTM is shown in Fig 15. Fig 15 (a) shows that the LSTM with full precision converges much faster with the lowest loss compared to BC, TC, and QC. However, validation result shows good accuracy for sentiment analysis. In both cases, TC and QC provide better performance compared to BC. Fig. 16 shows the testing accuracy on IMDB dataset. The experimental result shows testing accuracies of 82.87%, 79.64%, 76.86%, and 76.25% for FP, QC, TC, and BC respectively. We have achieved around 2.00% less on testing accuracy with QC and around 4% less accuracy compared against TC. There is however a significant advantage in terms of computational time and energy. In addition, this type of compressed version of recurrent approaches is suitable for embedded and mobile applications.

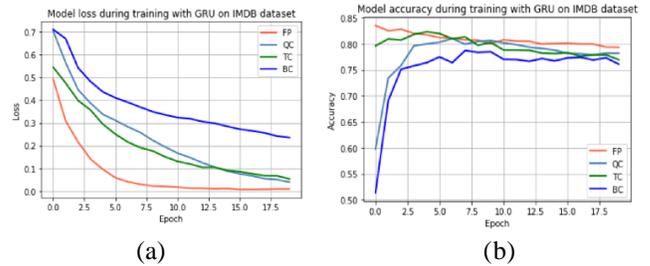

(a)            (b)

**Fig. 17.** Model loss and accuracy during training for GRU : (a) Loss and (b) Accuracy.

#### 2) Results with GRU
Training loss and validation accuracy for GRU are shown in Fig. 17 (a) and (b) respectively. In this experiment, GRU with quantization of BC, TC, and QC gives very good testing accuracy with respect to the full precision GRU. Testing accuracy is shown in Fig. 18

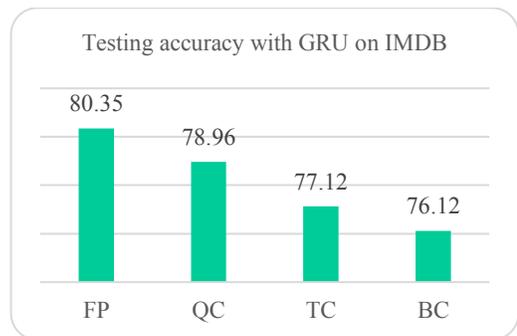

**Fig. 18**. Testing accuracy using GRU.

. However, LSTM provides overall better performance in most of the cases against GRU for sentiment analysis tasks.

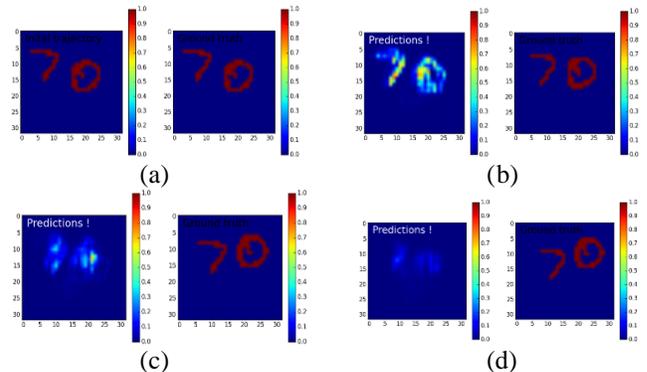

(a)            (b)

(c)            (d)

**Fig. 19.** (a) Output of Binary connection of ConvLSTM for actual trajectory of $7^{th}$ frame on left and ground truth on the right, (b) Predicted frame on the left and ground truth on the right for $8^{th}$ number frame, (c) Predicted frame on the left and ground truth on the right for 9th number frame and (d) Prediction result for 10th frame.

## B. Movie frames prediction

We have tested the performance of quantized ConvLSTM for object states prediction from the input video frames. We have implemented ConvLSTM with different quantization methods including BC, TC, QC, and ConvLSTM with full precision, which is tested on the moving MNIST dataset. There are 15 frames in total of the input moving MNIST dataset where seven frames are used for training. After training successfully, we have tried to generate posterior frames from frame number 8.

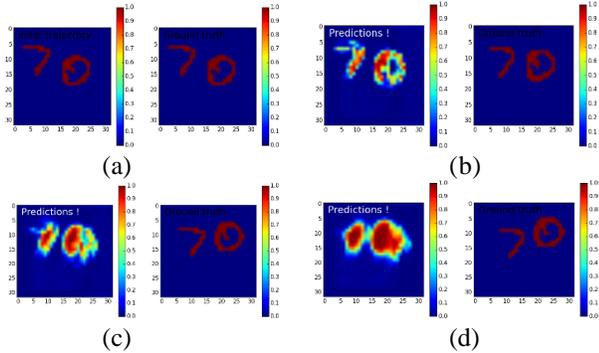

**Fig.20.** (a) Output of ternary connection of ConvLSTM for actual trajectory of 7th frame on left and ground truth on the right, (b) Predicted frame on the left and ground truth on the right for 8th number frame, (c) Predicted frame on the left and ground truth on the right for 9th number frame and (d) Prediction result for 10th frame.

The experiment illustrates promising results for video frame prediction on the moving MNIST dataset. In this implementation, we have applied 50 epochs for training. The following figure shows the predicted frames with BC ConvLSTM. Fig 19 (a) shows the initial trajectory and ground truth which is 7th frame. The prediction and ground truth of 8th, 9th, and 10th frames are shown in Fig. 19 (b), (c), and (d) respectively. The results for TC and QC are shown in Fig. 20 and 21 respectively which demonstrates the qualitative performance of ConvLSTM. The outputs of ConvLSTM with full precision are shown in Fig. 22. The experimental result shows good reconstruction compare to BC, TC, and QC. If we observe Fig. 22(d) then reconstruction of 10th frame is much more better than others.

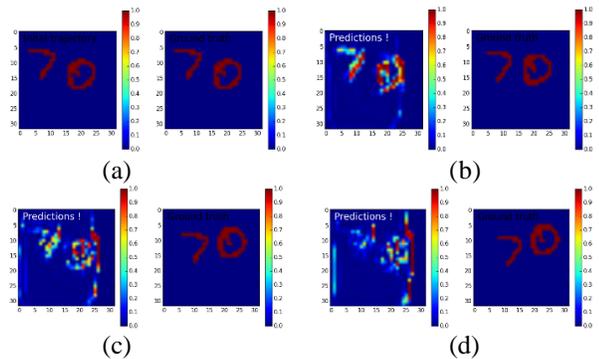

**Fig. 21.** (a) Output of quaternary connection of ConvLSTM for actual trajectory of 7th frame on left and ground truth on the right, (b) Predicted frame on the left and ground truth on the right for 8th number frame, (c) Predicted frame on the left and ground truth on the right for 9th number frame and (d) Prediction result for 10th frame.

For analysis the performance of ConvLSTM on moving MNIST experiment, we have calculated the MSE between input frames and predicted frames. In the following equation, *I* is the input frame and *K* is the predicted frame:

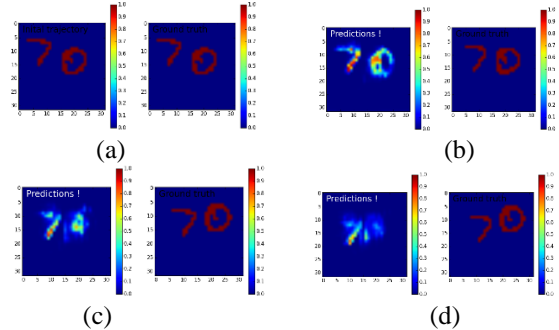

**Fig. 22.** (a) Output of full precision of ConvLSTM for actual trajectory of 7th frame on left and ground truth on the right, (b) Predicted frame on the left and ground truth on the right for 8th number frame, (c) Predicted frame on the left and ground truth on the right for 9th number frame and (d) Prediction result for 10th frame.

$$MSE = \frac{1}{m\,n}\sum_{i=0}^{m-1}\sum_{j=0}^{n-1}[I(i,j) - K(i,j)]^2 \qquad (25)$$

The following figure is showing the MSE for moving MNIST dataset where x-axis shows the number of frames and y-axis shows the MSE with respect to the frame predicted.

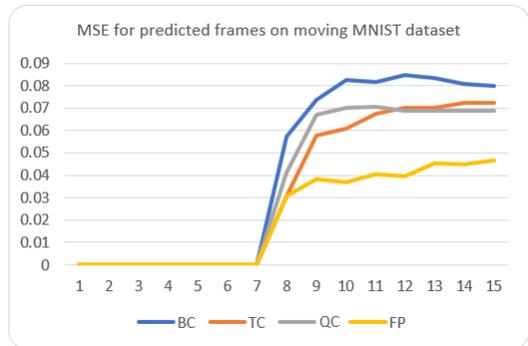

**Fig. 23.** MSE errors for moving MNIST frame prediction.

According to Fig. 23, it can be observed that the full precision ConvLSTM shows better performance in term of MSE compared to BC, TC, and QC. However, TC and QC also show promising results on the frames prediction task.

## V. CONCLUSION

In this work, we have proposed efficient quantization approaches for Recurrent Neural Networks (RNNs) including Long Short Term Memory (LSMT), Gated Recurrent Units (GRU), and Convolutional LSTM (ConvLSTM). The adaptive thresholding methods are proposed based on the basic statistics of the weights of a layer. We have also investigated the performance of approximate normal and uniform distribution of quantized weights for Binary Connect (BC), Ternary Connect (TC), and Quaternary Connect (QC) techniques. The empirical results show that the normal distribution shows better performance against uniform distribution with quantized weights. These proposed quantization methods are tested for sentiment analysis and movie frame generation on moving MNIST dataset. The results show promising performance against full precision for LSTM, GRU, and ConvLSTM. It is noted that this is the first-time experimental evaluation of the performance of the quantized ConvLSTM approach for movie frame generation. In the future, we would like to evaluate the performance of quantized ConvLSTM for more complex datasets.

ACKNOWLEDGMENT

This work was performed under the auspices of the U.S. Department of Energy by Lawrence Livermore National Laboratory under Contract DE-AC52-07NA27344 (LLNL-CONF-745577). Funding provided by ASC Beyond Moore's Law Program. Experiments were performed at the Livermore Computing facility resources.